# Low-rank MERA Decomposition of 3D OCT Image Volume for Compression, Denoising, Feature Extraction and Diagnosis of Retinal Diseases

Ivica Kopriva, *Senior Member, IEEE*, Fei Shi, Marija Štanfel, Qin Peng, Jianhao Bao, Gaowei Li, and Xinjian Chen

***Abstract*—Optical coherence tomography (OCT) is a non-invasive imaging modality used for visualization of retinal tissue. However, OCT volumes are affected by speckle noise, which degrades image contrast and obscures fine structural details. Furthermore, modern clinical practice requires large field-of-view, high-resolution OCT volumes, whose storage and transmission are demanding for archiving and teleophthalmology applications. To address aforementioned challenges, we propose a low-rank multi-scale entanglement renormalization ansatz (MERA) tensor network approximation of 3D OCT image volume, where the image volume is represented as contractions of a top core factor encoding global features and multiple interacting low-rank orthogonal or semi-orthogonal factors modeling inter- and intra-B scan correlations. This representation jointly enables compression, denoising, and feature extraction. Unlike deep learning approaches, the proposed approach is training-free and processes each OCT volume independently. We evaluate the proposed framework on OCT image volumes acquired by two different scanners and demonstrate improved visual quality and reduced retinal layer segmentation errors in comparison with several competing methods. We also extract MERA features from two clinical OCT volume datasets. They are subsequently classified using a Grassmann-manifold nearest-subspace classifier. With the subspace dimension as its only hyperparameter, the classifier is theoretically shown to be robust in small-sample clinical scenario. We demonstrate competitive accuracy of classification of retinal diseases in these small-sample settings, where deep learning methods are often limited by the scarcity of labeled data. The proposed structure-aware model-driven representation learning for high-dimensional data under limited supervision provides good generalization performance while maintaining high interpretability.**

## I. INTRODUCTION

OPTICAL coherence tomography (OCT) is a non-invasive imaging modality widely used for acquiring high-resolution cross-sectional images of the retina at micro-meter scale and plays a crucial role in the diagnosis of retinal diseases [1]. OCT exploits the coherence properties of light backscattered from biological tissue to reconstruct its internal microstructure [2], [3]. However, the same coherence properties inherently give rise to speckle noise, which is one of the principal factors degrading OCT image quality [2]. Speckle noise reduces image contrast and obscures fine structural details, making the boundaries between retinal layers and other morphological structures more difficult to delineate [4], [5], [6]. As a result, it compromises the accuracy of clinical interpretation and quantitative image analysis. Consequently, effective methods for speckle suppression remain essential for improving OCT image quality and facilitating reliable retinal disease diagnosis.

Image priors play a fundamental role in image denoising. According to the manner in which these priors are incorporated, denoising methods can be broadly classified into two categories: model-driven and learning-based approaches. Model-driven methods formulate image restoration as an optimization problem regularized by explicitly defined image priors, such as sparsity [7], nonlocal self-similarity [8], low-rankness [9], [10], and smoothness [11]. In contrast, learning-based methods learn a mapping from noisy to clean images using large collections of paired training data [12], [13]. While learning-based approaches have achieved remarkable denoising performance, their success depends heavily on the availability of representative annotated

This work was supported by the Croatian Science Foundation Grant HRZZ-IP-2022-10-6403, by the National Natural Science Foundation of China (62271337). and by the bilateral Croatian-Chinese Scientific and Technological Cooperation Project "Tensor models and tensor networks for compression, de-speckling, feature extraction and diagnosis of retinal diseases from 3D OCT images". *(Corresponding authors: Fei Shi and Xinjian Chen)*.

Ivica Kopriva is with Division of Computing and Data Science, Rudjer Boskovich Institute, Zagreb,Croatia (e-mail: ikopriva@irb.hr).

Fei Shi and Gaowei Li are with the School of Electronics and Information Engineering, Soochow University, Suzhou, China, and the Jiangsu Key Laboratory of Artificial Intelligence and Medical Imaging, Soochow University, Suzhou, China (e-mail: gwi7658@163.com, shifei@suda.edu.cn).

Xinjian Chen is with Health Management Center, Sichuan Provincial People's Hospital, School of Medicine, University of Electronic Science and Technology of China, Chengdu, China (e-mail: xjchen@uestc.edu.cn).

Marija Štanfel is with Department of Ophthalmology, University Hospital Center Zagreb, Zagreb, Croatia (e-mail: stanfelmstanfel@gmail.com).

Qing Peng and Jianhao Bai are with Department of Ophthalmology, Shanghai Tenth People's Hospital, School of Medicine, Tongji University, Shanghai, People's Republic of China (e-mail: pengqing@tongji.edu.cn, jianhao0701@gmail.com).

datasets and substantial computational resources for training. In contrast, model-driven methods are training-free, more interpretable and often exhibit stronger generalization, particularly in applications where only limited training data are available.

To facilitate comprehensive clinical assessment, ophthalmologists increasingly require large-field-of-view, high-spatial-resolution OCT image volumes [14]. However, the storage and transmission of such data impose a major burden on clinical archiving systems, teleophthalmology platforms, and large-scale screening programs, particularly as OCT datasets continue to grow in size [15]. This challenge is especially relevant for the diagnosis and longitudinal monitoring of retinal diseases, such as age-related macular degeneration [17]. Consequently, there is a growing need for efficient 3D OCT image compression methods that achieve high compression ratios while preserving image quality to support accurate clinical interpretation and quantitative image analysis.

As emphasized in [16], real-world medical datasets often exhibit a long-tailed class distribution, in which a small number of classes account for the majority of the available samples, whereas many classes are represented by only a limited number of instances. This imbalance is particularly pronounced for rare diseases, where collecting sufficiently large annotated datasets is inherently difficult [17]. Although deep learning has achieved remarkable success in medical image classification [18], [19], including OCT image classification [6], [20], [21], its performance and generalization capability often deteriorate in small-sample settings because of limited supervision and the increased risk of overfitting. Consequently, there remains a strong need for model-driven approaches that can deliver reliable performance in limited supervision scenario.

In the light of aforementioned challenges, an important question arises: Can structure-aware model-driven framework be devised for compression, denoising, feature extraction, and retinal disease diagnosis from OCT image volumes that processes each volume independently and generalizes reliably under limited supervision?

In this paper, we exploit the powerful multi-scale representation capability of the MERA tensor network [21], [22] to develop a compression ratio (CR)-constrained low-rank approximation of 3D OCT image volumes. By explicitly imposing a target CR, the approximation rank is directly controlled, enabling a flexible trade-off between compression efficiency and reconstruction fidelity. Owing to the inherent denoising effect of low-rank approximation, the reconstructed volumes often exhibit higher quality than the original noisy acquisitions, as demonstrated by reduced retinal layer segmentation errors and improved image quality assessed by expert ophthalmologists. Moreover, the proposed representation naturally provides a compact feature extractor by interpreting the top core factor of the MERA network as a feature representation. Specifically, the OCT volume is factorized into contractions of a top core and multiple interacting low-rank orthogonal or semi-orthogonal factors that capture complex intra- and inter-B-scan correlations across multiple spatial scales. Consequently, the top core benefits from the hierarchical structure of the MERA network and provides a compact feature representation that aggregates multi-scale information from the entire OCT volume, comparable to the role of the core tensor in the Tucker model [23].

Deep learning methods often experience reduced generalization performance when only a limited number of labeled training samples are available, as is frequently the case for rare retinal diseases [16], [17]. In contrast, many types of (medical) image data possess substantially lower intrinsic dimensionality than that of their ambient feature space [24], motivating the representation of high-dimensional data by low-dimensional subspaces. Building upon this observation, we represent each disease class $c \in \{0, \ldots, C\}$ by a point on the Grassmann manifold (GM), corresponding to a $d$-dimensional subspace estimated from the $D$-dimensional MERA-extracted feature vectors of the training samples, where $d \ll D$. Disease classification is then performed using a nearest-subspace-classifier (NSC), which assigns a test feature vector to the class whose subspace yields the smallest projection distance. Consequently, the proposed classifier has only a single hyperparameter, namely the subspace dimension $d$. As demonstrated in Section V, classification with competitive accuracy can be achieved with small values of $d$ (even with $d$=1), allowing the class subspaces to be reliably estimated from only a few training samples using singular value decomposition.

The major contributions of our paper are as follows:

1) We develop a CR-constrained low-rank MERA approximation of 3D OCT image volumes that provides a unified framework for image compression, denoising, and multi-scale feature extraction. Unlike learning-based approaches, the proposed framework is model-driven, training-free and processes each OCT volume independently, making it well suited for applications with limited training data.

2) We demonstrate that the proposed low-rank MERA approximation improves OCT image at substantial compression. Specifically, retinal layer segmentation performed on reconstructed OCT volumes acquired with two different scanners yields lower segmentation errors at CR as high as 40, than segmentation performed on the original noisy images as well as on images denoised by several state-of-the-art methods. This highlights the strong representation capability of the MERA tensor network.

3) We validate the clinical usefulness of the proposed framework through expert image quality assessment. Expert image quality metrics (EIQMs) remain competitive with those obtained using state-of-the-art denoising methods, even at substantially higher compression ratios, indicating that the reconstructed OCT volumes are suitable for clinical archiving, teleophthalmology, and visual assessment.

4) We introduce a GM-NSC operating on MERA-extracted features for diagnosis of retinal diseases in small-sample settings. Experimental results on two clinical OCT datasets [25], [26] show that the proposed classifier consistently outperforms support vector machine (SVM) and neural network (NN) classifiers optimized by hyperparameter tuning while possessing up to three orders of magnitude lower computational complexity in the training phase. Despite its simplicity, its classification performance is also comparable to that of deep network baselines, while preserving the interpretability and data efficiency of the proposed model-driven framework.

The remainder of this article is organized as follows. Section II reviews related work. Section III introduces notation and theoretical preliminaries of the MERA tensor network decomposition. Section IV presents the proposed CR-constrained low-rank MERA framework for compression, denoising, and feature extraction, as well as GM-NSC for classification of retinal diseases in small-sample settings scenario. Section V reports the experimental results and analysis, and Section VI concludes the article.

## II. Background and Related Work

In the most general sense, the OCT image denoising methods can be divided into two groups [27]: 1) denoising during image acquisition; 2) denoising after the image is captured. Methods from the first group are also known as hardware-based. They simultaneously capture and average different realizations of the same image [28]. Although effective, these methods demand complex and costly image acquisition systems.

The second group methods, a.k.a. software-based methods, are based on post-processing of acquired OCT image volume [29]. The filtering approaches use local statistics based on the knowledge of adjacent pixels [30]. They are known not to preserve the image details. The sparsity-based methods manipulate sparse coefficients extracted from OCT images using either fixed dictionaries [31] or learned dictionaries [32]. To learn a good dictionary a high signal-to-noise ratio is required. Performance of these methods is maximal when work with small patches (8×8 pixels) and small step size (1 pixel) [10], which makes them computationally demanding. A strong alternative to sparsity prior is low-rank prior used for OCT image denoising in low-rank matrix approximation [33], and low-rank tensor approximation [4], [10], [34]. As opposed to matrix rank, the tensor rank depends on tensor model which influences the quality of approximation of OCT image volume. In particular, the Tucker tensor model [35] and the tensor train (TT) model [36] were applied for OCT image volume denoising and compression in [10], while the tensor ring (TR) model [37] was used for OCT image volume denoising in [4]. The TT model represents higher order tensor as the interconnected tensors of order at most 3 and it is, therefore, free from the curse of dimensionality. However, limitations of TT representation related to its ranks, [37], affect quality of approximation of 3D OCT image volume. These limitations motivated derivation of the TR model [37]. Recently, MERA tensor network was successfully applied to the problem of multi-view subspace clustering [38] and hyperspectral anomaly detection [39], exploiting its strong representation power which benefits from multiple interactions among orthogonal or semi-orthogonal (low-rank) factors. In case of 3D OCT image volume, they capture complex intra- and inter-B-scan correlations, see Section III. This motivated us to apply low-rank MERA approximation of 3D OCT image volume for compression, denoising and feature extraction.

In addition to aforementioned approaches, deep learning methods have also been proposed for denoising of OCT image volumes. Supervised deep learning methods require significant number of noisy and clean image pairs [41], [42]. As it is pointed out in [6], due to involuntary eye movements and the high time necessary to obtain multiple scans, acquiring the large number of clean B-scans is non-trivial issue. This problem can be reduced by using semi-supervised learning [42]. However, in general the training phase determines the success of most deep learning strategies. This represents motivation to develop training-free approach that can be applied to individual OCT image volumes, as well as to propose classifier that can be trained in a small-sample setting under limited supervision.

## III. Notations and Theoretical Preliminaries

### A. Notations

$N$-order tensor is represented as $\mathcal{X} \in \mathbb{R}^{I_1 \times I_2 \times \ldots \times I_N}$ with the elements $x_{i_1 i_2 \ldots i_N}$ where $i_1$=1, ..., $I_1$, $i_2$=1, ..., $I_2$, ..., $i_N$=1, ..., $I_N$. Each index is called way or mode and $\{I_n\}_{n=1}^{N}$ are corresponding dimensions. $\mathbb{R}$ stands for the real manifold. That is standard notation adopted for use in multiway analysis [43]. Matrices are denoted as $\mathbf{X} \in \mathbb{R}^{I_1 \times I_2}$ and vectors as $\mathbf{x} \in \mathbb{R}^{I_1}$, while italic font letter represents scalar, e.g. $x$.

**Definition 1** (Mode-$\mathcal{S}$ unfolding). Let $\mathcal{X} \in \mathbb{R}^{I_1 \times I_2 \times \ldots \times I_N}$ and let $\mathcal{S} \subseteq \{1,2,\ldots,N\}$. The mode-$\mathcal{S}$ unfolding of $\mathcal{X}$, denoted by $\mathbf{X}_{[\mathcal{S}]}$, is the matrix obtained by reshaping $\mathcal{X}$ so that modes in $\mathcal{S}$ form the row index and the remaining modes form the column index while preserving lexicographic ordering. This definition generalizes the standard mode-$n$ unfolding [43].

For any two tensors $\mathcal{A} \in \mathbb{R}^{I_1 \times I_2 \times J_1 \times J_2}$ and $\mathcal{B} \in \mathbb{R}^{J_1 \times J_2 \times L_1 \times L_2}$, their tensor contraction is represented by $\mathcal{C} \in \mathbb{R}^{I_1 \times I_2 \times L_1 \times L_2}$, that is, $\mathcal{C} = \mathcal{A} \times_{\{J_1, J_2\}} \mathcal{B}$, where $c_{i_1, i_2, l_1, l_2} = \sum_{j_1, j_2} a_{i_1, i_2, j_1, j_2} b_{j_1, j_2, l_1, l_2}$.

### B. MERA Decomposition Model

The three basic components of MERA decomposition are: isometry, disentangler, and layer [21], [22].

**Definition 2**. (Isometry). An isometry $\mathcal{W} \in \mathbb{R}^{I_1 \times I_2 \times \ldots \times I_K \times R}$ is denoted as a ($K$+1)-th order tensor with semi-orthonormal unfolding matrix $\mathbf{W} \in \mathbb{R}^{I_1 \ldots I_K \times R}$, e.g., $\mathbf{W}^{\mathrm{T}}\mathbf{W}=\mathbf{I}_R$, where $R$ is the outgoing dimension of an isometry and $K$ is the number of bottom legs.

**Definition 3** (Disentangler). A disentangler $\mathcal{U} \in \mathbb{R}^{I_1 \times I_2 \times I_1 \times I_2}$ is denoted as a 4-way tensor with a square orthogonal unfolding matrix $\mathbf{U} \in \mathbb{R}^{I_1 I_2 \times I_1 I_2}$, namely, $\mathbf{U}^{\mathrm{T}}\mathbf{U} = \mathbf{\Lambda}_{I_1 I_2}$, and $\mathbf{\Lambda}$ represents a diagonal matrix.

**Definition 4** (Layer). A layer $\mathcal{C}$ is denoted as a result of tensor contraction between isometries and disantanglers in the same layer. For example, a 7-th order layer $\mathcal{C}$ can be constructed by two isometries $\mathcal{W}_1 \in \mathbb{R}^{I_1 \times I_2 \times R_1}$ and $\mathcal{W}_2 \in \mathbb{R}^{I_3 \times I_4 \times I_5 \times R_2}$, and one disantangler $\mathcal{U} \in \mathbb{R}^{I_2 \times I_3 \times I_2 \times I_3}$ as follows: $\mathcal{C} = \mathcal{U}_1 \times_{\{I_2\}} \mathcal{W}_1 \times_{\{I_3\}} \mathcal{W}_2 \in \mathbb{R}^{I_1 \times I_2 \times I_3 \times I_4 \times I_5 \times R_1 \times R_2}$.

**Definition 5** (MERA decomposition). For a $D$-th order tensor $\mathcal{Y} \in \mathbb{R}^{I_1 \times I_2 \times \ldots \times I_D}$, its MERA representation can be composed by different layers and one top core which can be recursively obtained by $\mathcal{G}^{l-1} = \mathcal{G}^{l} \times_{\{R_1^l, \ldots, R_{p_l}^l\}} \mathcal{C}_l$, where $R_{p_l}^l$, $p_l$ = 1, ..., $L$ are the MERA ranks; $p_l$ is the number of isometries in the $l$-th layer, and $L$ is the total number of layers. For $L$=1 network, shown in Fig. 1, we have $\mathcal{G}^0 = \mathcal{Y}$, and $\mathcal{G}^1 = \hat{\mathcal{Y}} = \mathcal{C} \times_{\{R_1,R_2\}} \mathbf{G} \in \mathbb{R}^{I_1 \times I_2 \times I_3 \times I_4 \times I_5}$, where $\mathbf{G}$ is the top core. For simplicity, we use $\mathcal{Y} = f(\mathbf{G}, \mathcal{W}_1, \mathcal{W}_2, \mathcal{U}_1)$ to represent the one-layer MERA decomposition.

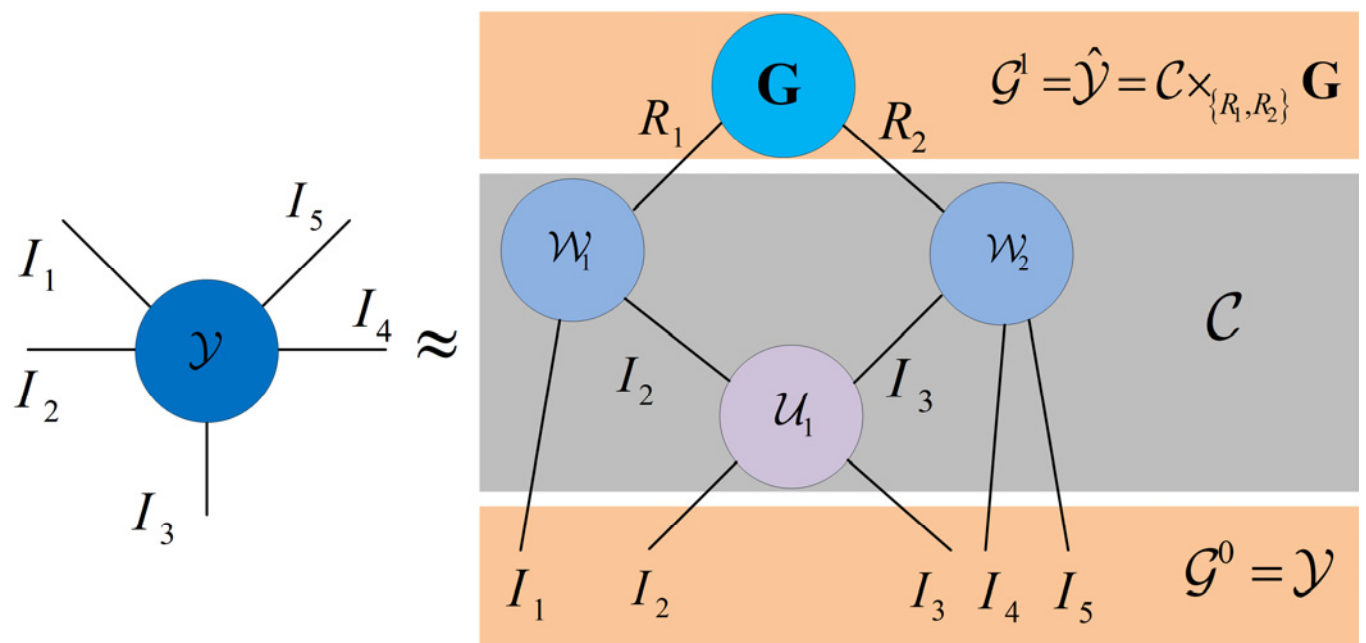

Fig. 1. Graphical illustration of MERA decomposition for a 5-th order tensor.

### C. Low-Rank MERA Approximation

From [38], [39] we reproduce the low-rank MERA approximation algorithm for a 5-th order tensor $\mathcal{Y} \in \mathbb{R}^{I_1 \times .. \times I_5}$. The low-rank MERA approximation consists in finding the MERA factors with predefined MERA ranks $R_1$=$R_2$=$R$, which is formulated as:

$$\min_{\mathcal{U}_1, \mathcal{W}_1, \mathcal{W}_2, \mathbf{G}} \tfrac{1}{2} \left\| \mathcal{Y} - f(\mathbf{G}, \mathcal{W}_1, \mathcal{W}_2, \mathcal{U}_1) \right\|_F^2 . \tag{1}$$

The problem is solved in alternating manner.

*Update* $\mathcal{U}_1$: Fixing $\mathcal{W}_1$, $\mathcal{W}_2$, and $\mathbf{G}$, the subproblem of $\mathcal{U}_1$ in matrix form becomes:

$$\min_{\mathbf{U}_1} \tfrac{1}{2} \left\| \mathbf{Y}_{[2,3]} - \mathbf{U}_1 \mathbf{M}_u \right\|_F^2 \quad \text{s.t.}\ \mathbf{U}_1^{\mathrm{T}} \mathbf{U}_1 = \mathbf{\Lambda}_{I_2 I_3}, \tag{2}$$

where $\mathbf{Y}_{[2,3]}$ and $\mathbf{M}_u$ are the mode–[2, 3] unfolding matrices of $\mathcal{Y}$ and $\mathcal{M}_u$, respectively. $\mathcal{M}_u$ is tensor contraction outcome of other variables, $\mathcal{M}_u = \mathbf{G} \times_{\{R_1\}} \mathcal{W}_1 \times_{\{R_2\}} \mathcal{W}_2 \in \mathbb{R}^{I_1 \times I_2 \times I_3 \times I_4 \times I_5}$; $\mathbf{U}_1$ is the mode-[1,2] unfolding matrix of $\mathcal{U}_1 \in \mathbb{R}^{I_2 \times I_3 \times I_2 \times I_3}$. The optimization problem of (1) is equivalent to the orthogonal Procrustes problem:

$$\max_{\mathbf{U}_1} \operatorname{trace}\left( (\mathbf{U}_1)^{\mathrm{T}} \mathbf{Y}_{[2,3]} (\mathbf{M}_u)^{\mathrm{T}} \right). \tag{3}$$

The update of $\mathcal{U}_1$ is obtained from (3) as:

$$\mathcal{U}_1 = \operatorname{fold}_{1,2}\left( \mathbf{S}_u \mathbf{D}_u^{\mathrm{T}} \right), \tag{4}$$

where $\mathbf{S}_u$ and $\mathbf{D}_u$ are left and right singular vector matrices of the matrix $\mathbf{Y}_{[2,3]}(\mathbf{M}_u)^{\mathrm{T}}$.

*Update* $\mathcal{W}_1$, $\mathcal{W}_2$: The update of $\mathcal{W}_p$, $p$=1, 2 can be obtained analogously to $\mathcal{U}_1$:

$$\mathcal{W}_p = \operatorname*{arg\,min}_{\mathcal{W}_p : \mathbf{W}_p^{\mathrm{T}} \mathbf{W}_p = \mathbf{I}_{R_p}} \tfrac{1}{2} \left\| \mathcal{Y} - f(\mathbf{G}, \mathcal{W}_1, \mathcal{W}_2, \mathcal{U}_1) \right\|_F^2 . \tag{5}$$

To be more specific, the subproblem of $\mathcal{W}_1$ in matrix form becomes:

$$\min_{\mathbf{W}_1} \tfrac{1}{2} \left\| \mathbf{Y}_{[1,2]} - \mathbf{W}_1 \mathbf{M}_w \right\|_F^2 \quad \text{s.t.}\ \mathbf{W}_1^{\mathrm{T}} \mathbf{W}_1 = \mathbf{I}_{R_1} \tag{6}$$

where $\mathbf{Y}_{[1,2]}$ is the mode -[1, 2] unfolding matrix of $\mathcal{Y}$, and $\mathbf{M}_w$ is the mode [1] unfolding matrix of $\mathcal{M}_{w_1}$. $\mathcal{M}_{w_1}$ is tensor contraction outcome of other variables, $\mathcal{M}_{w_1} = \mathbf{G} \times_{\{R_2\}} \mathcal{W}_2 \in \mathbb{R}^{R_1 \times I_3 \times I_4 \times I_5}$; $\mathbf{W}_1$ is the mode-[1,2] unfolding matrix of $\mathcal{W}_1 \in \mathbb{R}^{I_1 \times I_2 \times R_1}$. Solution of (6) is obtained analogously to the solution of (2)-(4). The subproblem of $\mathcal{W}_2$ in matrix form becomes:

$$\min_{\mathbf{W}_2} \tfrac{1}{2} \left\| \mathbf{Y}_{[3,4,5]} - \mathbf{W}_2 \mathbf{M}_w \right\|_F^2 \quad \text{s.t.}\ \mathbf{W}_2^{\mathrm{T}} \mathbf{W}_2 = \mathbf{I}_{R_2} \tag{7}$$

where $\mathbf{Y}_{[3,4,5]}$ is the mode -[3, 4,5] unfolding matrix of $\mathcal{Y}$, and $\mathbf{M}_w$ is the mode [1] unfolding matrix of $\mathcal{M}_{w_2}$. $\mathcal{M}_{w_2}$ is tensor contraction outcome of other variables, $\mathcal{M}_{w_2} = \mathbf{G} \times_{\{R_1\}} \mathcal{W}_1 \in \mathbb{R}^{R_2 \times I_1 \times I_2}$; $\mathbf{W}_2$ is the mode-[1,2,3] unfolding matrix of $\mathcal{W}_2 \in \mathbb{R}^{I_3 \times I_4 \times I_5 \times R_2}$. Solution of (7) is obtained analogously to the solution of (2)-(4). The subproblem of $\mathbf{G}$ can be rewritten as:

$$\min_{\mathbf{G}} \tfrac{1}{2} \left\| \mathcal{Y} - \mathcal{C} \times_{\{R_1,R_2\}} \mathbf{G} \right\|_F^2 \tag{8}$$

where $\mathcal{C}$ represents a 7-th order layer in Definition 4. Then, due to orthogonality of disentangler $\mathcal{U}_1$ and orthonormality of isometries $\mathcal{W}_1$ and $\mathcal{W}_2$ the update of $\mathbf{G}$ can be obtained by:

$$\mathbf{G} = \mathcal{Y} \times_{\{I_1, I_2, I_3, I_4, I_5\}} \mathcal{C} . \tag{9}$$

## IV. Proposed Methodology

### A. Low-Rank MERA Approximation for Compression, Denoising and Feature Extraction

Let $\mathcal{X} \in \mathbb{R}^{I_1 \times I_2 \times I_3}$ represent a 3D OCT image volume with $I_1$ rows, $I_2$ columns and $I_3$ B-scans [10]. Here we present application of the low-rank MERA approximation of 3D OCT image volume of a retina for compression, denoising and feature extraction. We transform 3D OCT image tensor $\mathcal{X}$ to 5D tensor $\mathcal{Y}$:

$$\begin{aligned} &\mathcal{X} \in \mathbb{R}^{I_1 \times I_2 \times I_3} \rightarrow \mathcal{Y} \in \mathbb{R}^{J_1 \times J_2 \times J_3 \times J_4 \times I_3} \\ &\text{s.t.}\ I_1 = J_1 \times J_2, I_2 = J_3 \times J_4 \end{aligned} \tag{10}$$

For pre-specified rank $R_1$=$R_2$=$R$ we perform the low-rank MERA approximation of $\mathcal{Y}$ which yields: top core $\mathbf{G} \in \mathbb{R}^{R \times R}$, disantangler $\mathcal{U}_1 \in \mathbb{R}^{J_2 \times J_3 \times J_2 \times J_3}$, isometry $\mathcal{W}_1 \in \mathbb{R}^{J_1 \times J_2 \times R}$, and isometry $\mathcal{W}_2 \in \mathbb{R}^{J_3 \times J_4 \times I_3 \times R}$. We define the compression ratio (CR) as a ratio between number of elements of data tensor $\mathcal{Y}$ and number of elements of its MERA decomposition:

$$CR = \frac{\overbrace{J_1 J_2}^{I_1} \overbrace{J_3 J_4}^{I_2} I_3}{\underbrace{J_1 J_2}_{I_1} R + \underbrace{J_3 J_4}_{I_2} I_3 R + \underbrace{J_1 J_2}_{I_1} \underbrace{J_3 J_4}_{I_2} + R^2} \tag{11}$$

Equation (11) enables us to calculate MERA rank [$R$ $R$] as a function of specified $CR$ and dimensions of data tensor $\mathcal{Y}$. The problem boils down to solving quadratic equation with the solutions:

$$R_{1,2} = \frac{-b \pm \sqrt{d}}{2a} \quad (12)$$

where: $a = CR$, $b = CR(I_1 + I_2 I_3)$, $c = (CR - I_3) I_1 I_2$, $d = b^2 - 4ac$. Because $R \in \mathbb{N}$ the final solution is obtained as:

$$R = round(R_1) \quad (13)$$

where *round* maps the $R_1$ towards nearest integer. The necessary condition for $R_1 > 0$ is:

$$CR < I_3 . \quad (14)$$

Hence, the MERA rank $[R\ R]$ is directly controlled by prespecified CR enabling a flexible trade-off between compression efficiency and reconstruction fidelity. The low-rank MERA approximation (1) to (9) computes the estimates of the factors of the MERA tensor network model:

$$(\mathcal{U}_1, \mathcal{W}_1, \mathcal{W}_2, \mathbf{G}) \quad (15)$$

which represents the compressed form of the original 3D OCT image volume $\mathcal{X}$. To restore the compressed version of the OCT image volume we first compute the layer tensor as:

$$\mathcal{C} = \mathcal{U}_1 \times_{\{I_2\}} \mathcal{W}_1 \times_{\{I_3\}} \mathcal{W}_2 . \quad (16)$$

The top core $\mathbf{G}$ is computed using (9). We now compute the estimate of $\mathcal{Y}$ as:

$$\hat{\mathcal{Y}} = \mathcal{C} \times_{\{R_1, R_2\}} \mathbf{G} . \quad (17)$$

Finally, we compute the low-rank MERA approximation of 3D OCT image volume $\mathcal{X}$ as:

$$\hat{\mathcal{X}} = reshape\left(\hat{\mathcal{Y}}\right) . \quad (18)$$

Owing to denoising effect of the low-rank approximation, $\hat{\mathcal{X}}$ can be used for segmentation of retinal layers and visual diagnosis, see Section V. It is evident from Fig. 1 that tope core $\mathbf{G}$ interconnects all three modes of the 3D OCT image volume. Thus, its elements represent global features that encode inter/intra-B scan information characteristic on retinal diseases. Furthermore, by predefined value of the CR we control the number of features which is $R^2$.

## B. Feature Extraction

We assume that features which characterize 3D OCT image volume with diagnosis specific information are contained in the top core $\mathbf{G}$. We now obtain the feature vector as:

$$\mathbf{g} \in \mathbb{R}^{R^2 \times 1} = \mathrm{vec}\left(|\mathbf{G}|\right) \quad (19)$$

where vec() denotes vectorization and $|\mathbf{G}|$ denotes absolute value applied entry-wise. Let us suppose the set of 3D OCT image volumes contains images from $C$+1 classes corresponding to healthy class and $C$ diseases of a retina. Thus, we can group the feature vectors (19) according to:

$$\left\{ \mathbf{G}_c := \left[ \mathbf{g}_c^1 \ldots \mathbf{g}_c^{N_c} \right] \in \mathbb{R}^{R^2 \times N_c} \right\}_{c=0}^{C} \quad (20)$$

where in (20) $N_c$ represents number of 3D OCT image volumes in class $c$. As emphasized before, when only a limited number of labeled training samples are available, as it is the case for rare retinal diseases [16], [17], deep learning experiences reduced generalization performance. Herein, we propose GM-NSC method for such scenarios.

## C. Grassmann-Manifold Nearest Subspace Classifier

Classification of MERA-extracted feature vectors $\{\mathbf{G}_c\}_{c=0}^{C}$ can be understood as a problem of clustering high-dimensional data with complex intrinsic relations and nonlinear manifold structure. This problem is traditionally addressed by kernel methods, such as SVM and NN. However, GM-based representation is legitimate candidate for addressing such problem, see [44] for clustering data from nonlinear manifolds, and [45] for cancer related pixel-wise classification of hyperspectral images. Let us assume that MERA-extracted feature vector sits in $D = R^2$ dimensional ambient space. The canonical union-of-subspaces (UoS) model asserts that data in $D$-dimensional ambient space can be represented by a union of $C$+1 low-dimensional subspaces [46]: $\mathcal{M}_C = \cup_{c=1}^{C} \mathcal{S}_c$, where $\mathcal{S}_c$ is a subspace in $\mathbb{R}^D$. The fact that in many cases data exhibit fewer degrees of freedom than their ambient domain, i.e., $\dim(\mathcal{S}_c) = d_c \ll D$, motivates the identification of low-dimensional structures within high-dimensional spaces [47]. We further assume $\{d_c = d\}_{c=1}^{C}$, and each $\mathcal{S}_c$ represents a point on the Grassmann manifold $\mathcal{G}(D,d)$, which denotes a set of all $d$-dimensional linear subspaces of $D$-dimensional Euclidean spaces [48]. We have to estimate representatives of Grassmann points $\left\{\mathbf{Y}_c \in \mathbb{R}^{Dxd}\right\}_{c=0}^{C}$. For that purpose, we form the training sets $\left\{\mathbf{G}_c \in \mathbb{R}^{D \times N_c}\right\}_{c=0}^{C}$ from which we estimate representatives of Grassmann points according to:

$$\begin{aligned} \mathbf{G}_c &= \mathbf{U}_c \boldsymbol{\Sigma}_c \mathbf{V}_c^{\mathrm{T}} \\ \mathbf{Y}_c &= \mathbf{U}_c(:,1:d) \quad c \in \{0,\ldots,C\} \end{aligned} \quad (21)$$

where $d$ in (21) represents the *unknown* subspace dimension. To classify unlabeled data point we use minimum of point-to-subspace distance criterion [49]. We first compute:

$$\left\{ \mathbf{G}_c \leftarrow \mathbf{G}_c - \underbrace{\left[\boldsymbol{\mu}_c \ldots \boldsymbol{\mu}_c\right]}_{N_c \text{ times}} \right\} \quad c \in \{0,\ldots,C\} \quad (22)$$

where:

$$\boldsymbol{\mu}_c = \frac{1}{N_c} \sum_{i=1}^{N_c} \mathbf{G}_c(:,i) \quad c \in \{0,\ldots,C\} . \quad (23)$$

By using (21), (22) and (23) we estimate representatives $\left\{\mathbf{Y}_c \in \mathbb{R}^{Dxd}\right\}_{c=0}^{C}$. We now assign label to new MERA feature vector $\mathbf{g}$ according to [49]:

$$l(\mathbf{g}) = \underset{c \in \{0,\ldots,C\}}{\arg\min} \left\| \tilde{\mathbf{g}}_c - \mathbf{Y}_c \mathbf{Y}_c^{\mathrm{T}} \tilde{\mathbf{g}}_c \right\|_2 \quad (24)$$

where $\tilde{\mathbf{g}}_c = \mathbf{g} - \boldsymbol{\mu}_c$. We summarize MERA-based GM-NSC approach in the Algorithm 1.

**Algorithm 1**. Low-rank MERA feature extraction and GM-NSC for diagnosis of retinal diseases from 3D OCT image volume.

**Input**: Sets $\left\{\left\{\mathcal{X}_c^t \in \mathbb{R}^{I_1 \times I_2 \times I_3}\right\}_{t=1}^{N_c}\right\}_{c=0}^{C}$ of 3D OCT image volumes of a retina belonging to $C$+1 classes, where $c$=0 represents healthy retina; Compression ratio *CR*.

**1.** Map each 3D OCT image tensor to 5D tensor according to (10).

**2.** Perform low-rank MERA approximation of (10) according to (1)-(9).

**3**. Extract feature vectors according to (19).

**4**. Group feature vectors according to (20).

**5**. Estimate Grassmann points according to (21).

**6**. Assign label (diagnosis) to test feature vector using the NSC (24).

**Output**: Diagnosis assigned to the specific test 3D OCT image volume of a retina.

### D. Generalization in a small-sample scenario

It is evident that proposed GM-NSC depends on only one hyperparameter, the subspace dimension $d$. The central question to answer is why GM-NSC is expected to generalize well in a small-sample dataset scenario? It is clear from (21) and (24) that accuracy of GM-NSC depends critically on how well the true, but unknown, $d$-dimensional subspaces bases $\left\{\mathbf{S}_c\right\}_{c=0}^{C}$ are approximated by their estimates $\left\{\mathbf{Y}_c\right\}_{c=0}^{C}$. The answer is provided by the Davis-Kahan sinΘ theorem [50], i.e., its variant derived for statisticians [51]. In our notation Frobenious norm of error between estimated and true basis is given as [51]:

$$\left\|\mathbf{Y}_c\mathbf{O}-\mathbf{S}_c\right\|_F \leq \frac{2\sqrt{2d}\,\|Noise\|_{op}}{\delta}. \tag{25}$$

In equation (25), $\mathbf{O}$ is orthonormal rotation matrix, $\|Noise\|_{op}$ represents power of random noise in data and it is proportional with $1/\sqrt{N_c}$ . $\delta$ represents eigenvalue gap defined as:

$$\delta = \min\left(\sigma_d - \sigma_{d+1}, \sigma_0 - \sigma_1\right)$$

where $\sigma_0 \geq \sigma_1 \geq \ldots \geq \sigma_d \geq \sigma_{d+1} \geq \ldots \geq \sigma_{N_c}$ are singular values in (21). In contemporary data analysis $\sigma_0$ is always much greater than the rest of singular values. Therefore, $\delta = \sigma_d - \sigma_{d+1}$. We can now write (25) as:

$$\left\|\mathbf{Y}_c\mathbf{O}-\mathbf{S}_c\right\|_F \leq \frac{2\sqrt{2}}{\sigma_d - \sigma_{d+1}}\sqrt{\frac{d}{N_c}}$$

i.e., for a fixed $N_c$:

$$\left\|\mathbf{Y}_c\mathbf{O}-\mathbf{S}_c\right\|_F \approx \sqrt{d}\ . \tag{26}$$

The error in subspace basis approximation can be significantly reduced if subspace dimension $d$ is small relative to $D$, a common scenario in high-dimensional data analysis. As can be seen in Section V, for age-related macular degeneration [25], and glaucoma grading [26], subspace dimension $d$ is never greater than 12, and it is quite often equal to 1. Thus, the basis approximation error bound can be kept small even if size of training set $N_c$ is small. In such a case label assignment criterion (25) will be reasonably accurate, i.e., generalization capability of GM-NSC classifier will remain good.

## V. Experimental Methods and Results

In this section we present results of CR-constrained low-rank MERA decomposition for compression, denoising and feature extraction with diagnosis of retinal diseases.

### A. Data and Code

Proposed algorithms for segmentation of retinal layers and improvement of visual quality were compared on two sets of 3D OCT image volumes acquired by Topcon 3D OCT-1000 scanner (Topcon, Tokyo, Japan)[1] and by the non-commercial in-house BV1000 3D OCT scanner (BigVision, Suzhou, China).[2] In the first set 3D OCT image tensor is comprised of $I_3$=64 B-scan images with the size of $I_1$=480 × $I_2$=512 pixels. The scanning range covered a retinal area of 6 mm × 6 mm. First dimension refers to width of each B-scan. These images were used previously for the study for optical intensity analysis, where they were segmented into retinal layers [52], [10]. The second set is comprised of twenty macular-centered 3D OCT images of normal eyes with optical nerve included. 3D OCT image tensor is comprised of $I_3$=100 B-scan images with the size of $I_1$=1000 × $I_2$=512 pixels. The scanning range covered area of 12 mm×9 mm. The BVS scanner yields images of significantly lower quality which represents challenge for segmentation of retinal layers and visual quality enhancement. Data used for feature extraction and diagnosis of retinal diseases are available at links presented in [25] and [26]. CR-constrained low-rank MERA approximation, feature extraction and retinal disease diagnosis were implemented in MATLAB (version 2024b) under Windows 11 operating system running on 512 GB of RAM and AMD Ryzen PRO 5976WX processor with 32-Cores and 3.60 GHz clock. Deep learning models were trained on an NVIDIA RTX 5090 GPU with 32 GB of memory. The batch size was set to 4 and training was performed for 100 epochs. MATLAB and PyThorch code with demos on compression, feature extraction and diagnosis of retinal diseases are publicly available at https://github.com/ikopriva/MERA_3DOCT.

### B. Segmentation of Retinal Layers and Improvement of Visual Quality

MERA results were compared with fully connected tensor network (FCTN) [53]; low multilinear rank (MLR)- [10], [34], low TT approximations [10], [36], median filtering (MF) [10], and bilateral filtering (BF) [10]. Many other compression methods analyzed in [10] demonstrated lower performance. Therefore, we did not use them in the current study. Prior to segmentation of retinal layers, B-scan alignment was performed to compensate for involuntary eye movement [54]. For segmentation we used the algorithm presented in [54]. Fig. 2 shows segmentation error vs. CR for two scanners in percentage of thickness of a retina. In case of TOPCON scanner MERA-approximation achieved segmentation error as low as 1.88% for CR=4, and 1.87% for CR=10. At CR=40, only a no-compression

[1] https://pan.baidu.com/s/1edkG7k8W3Wkjhq8vffCYng (password: frpa).

[2] https://www.bigvisiontech.com/services/ bv1000

MF method slightly outperformed MERA. In case of BVS scanner MERA yields segmentation error below 4% even when CR=50, and close to 3% for CR≤30 comparing fairly with no-compression MF and BF methods. Achieved results are important for machine learning-based diagnosis using segmented retina layers [6]. Achieved 1.87% error with TOPCON scanner is important for layer-thickness based prediction of cardiovascular diseases [55].

To asses visual quality three ophthalmologists graded three B-scans from three 3D OCT volumes for each scanner. Expert image quality metric (EIQM), shown in Fig. 3, is presented as mean ± standard deviation of grades between 1 (the worse) and 5 (the best). In case of TOPCON scanner, MERA achieved EIQM compares fairly with EIQM of original image as well as of the low-rank TT, MF and BF enhanced images for CR≤30. Thereby, MF and BF are no-compression methods. It is better than FCTN and slightly worse than low-rank MLR enhanced image and original image. The reason is that TOPCON scanner acquired OCT image volumes which are already of high-quality. This, compression artifacts have noticeable effects on image quality. In case of BVS scanner low-rank MERA approximation together with low-rank MLR approximation highly improves visual image quality for CR≤40. Achieved results are important for clinical data archiving and remote consultation.

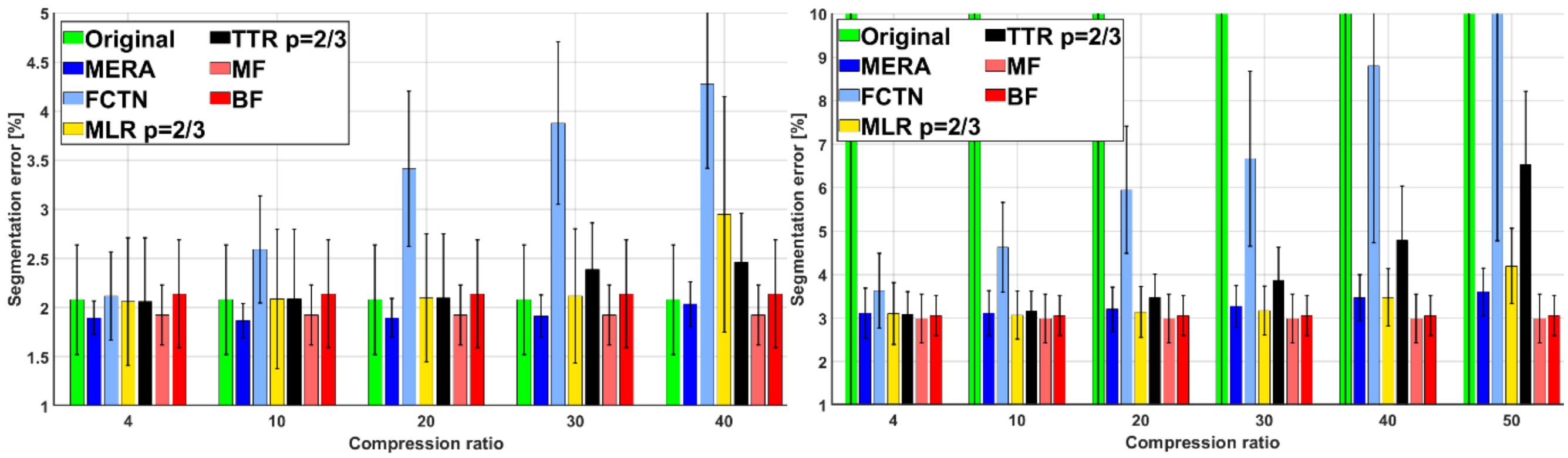

Fig. 2. Segmentation error of retinal layers in percentage of thickness of a retina as function of compression ratio. Left: TOPCON scanner. Right: BVS scanner. p=2/3 implies that Schatten 2/3 norm was used in low-rank approximation.

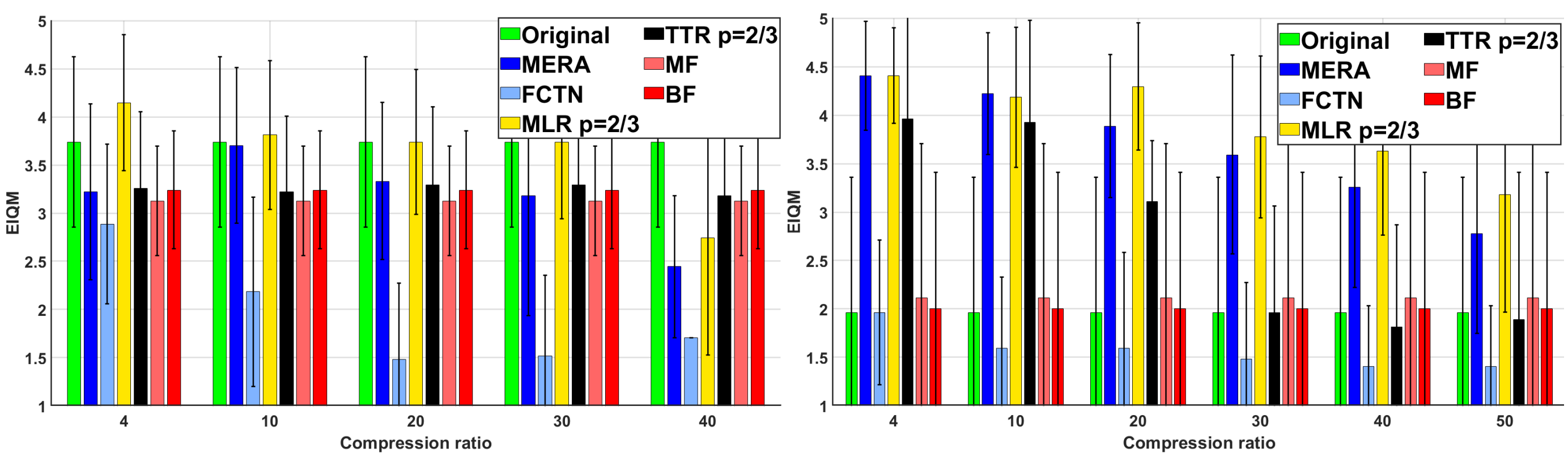

Fig. 3. Expert-based image quality metrics (EIQM). Left: TOPCON scanner. Right: BVS scanner. p=2/3 implies that Schatten 2/3 norm was used in low-rank approximation.

### *C. Feature Extraction and Diagnosis of Retinal Diseases*

#### *1) Age-related Macular Degeneration*

Corresponding dataset contains 114 normal/control cases and 269 AMD/disease cases [25]. To train the GM-NSC we first validate assumption on low-dimensionality of representation of extracted MERA features. For CR $\in$ {1.6, 2, 4, 10, 20, 40} we generated 100 random partitions such that 70% of samples are used for training and validation and 30% for testing. The subspace dimension was validated in the range $d \in$ {1, …, 40}. Fig. 4 shows accuracy as a function of subspace dimension $d$ with CR as parameter. For CR values in order given above maximum of accuracy occurs in respective order as $d \in$ {12, 11, 10, 6, 6, 6}. Thus, assumption on low-dimensionality of representation of high-dimensional MERA features holds. Since approximation error of the subspace basis is according to (26) proportional to $\sqrt{d}$ , it increases from 2.45 to 3.46. Note that values of the ambient space dimension that correspond with CR values given above are $D \in$ {98956, 62001, 4884, 2116, 400, 64}.

We now compare GM-NSC with the SVM, NN and ResNet18 deep classifier as baseline. Thereby, GM-NSC, SVM and NN classifiers were tested on 100 random partitions, while ResNet18 was tested on 30 partitions. The SVM classifier is trained using `fitcsvm` command with: (a) `OptimizeHyperparameters` set to all. It explores kernels (`linear`, `RBF` and `polynomial`), standardization (`false` and `true`) and kernel parameters; (b) `OptimizeHyperparameters` set to auto. For selected kernels their parameters are tuned together with standardization (`false` and `true`); (c) linear kernel with non-standardized data. Surprisingly, linear SVM yielded highest accuracy (ACC) indicating linear separability of feature vectors extracted by MERA tensor network. Thus, we selected linear SVM on non-standardized data for further comparative performance analysis with results presented in Table I.

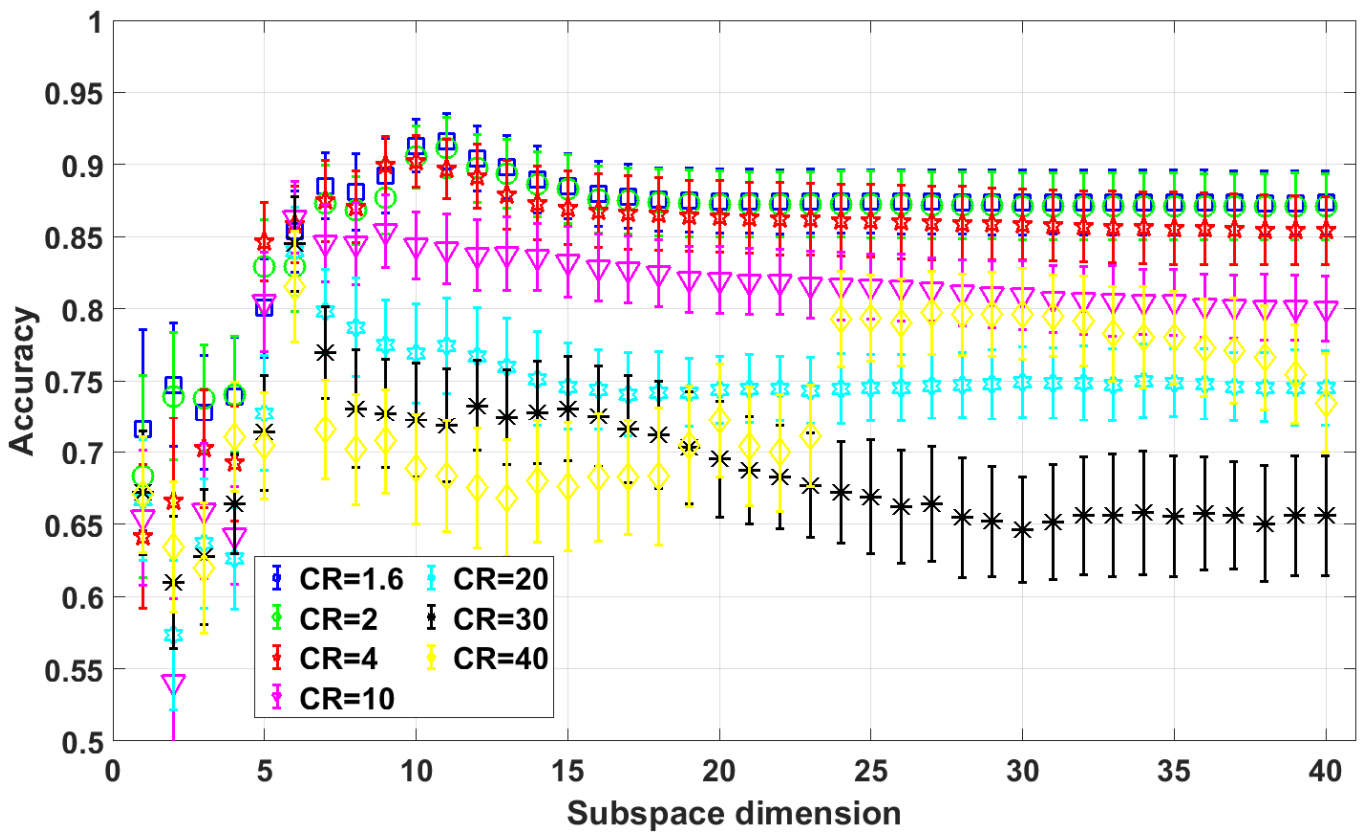

Fig. 4. Accuracy (mean ± standard deviation) vs. subspace dimension with compression ratio (CR) as parameter.

TABLE I
RESULTS OF OPTIMIZED SVM, NN AND GM-NSC CLASSIFIERS ON MERA-EXTRACTED FEATURE VECTORS FOR SELECTED CR. THE BEST RESULT IS IN BOLD.

| CR | Classifier | ACC [%] | CPU [s] |
|---|---|---|---|
| 1.6 | NN | 85.42±6.78 | 14079 |
| | SVM | 88.51±3.19 | 5089 |
| | **GM-NSC** | **91.61±1.94** | **41.45** |
| 2 | NN | 85.07±7.01 | 8141 |
| | SVM | 89.25±2.50 | 573.2 |
| | **GM-NSC** | **91.18±2.11** | **18.55** |
| 4 | NN | 87.06±3.34 | 1850.6 |
| | SVM | 89.18±2.44 | 29.92 |
| | **GM-NSC** | **90.21±1.78** | **1.85** |
| 10 | NN | 83.35±4.86 | 419.86 |
| | SVM | **88.40±2.43** | 1.31 |
| | GM-NSC | 86.33±2.49 | **0.10** |
| 20 | NN | 77.30±6.50 | 143.65 |
| | SVM | **87.31±2.44** | 0.127 |
| | GM-NSC | 83.96±3.06 | **0.013** |
| 40 | NN | 81.14±3.68 | 200.74 |
| | SVM | **83.27±2.76** | 0.059 |
| | GM-NSC | 81.54±3.83 | **0.004** |
| | ResNet18 | 92.35±3.30 | 5394 |

NN classifier was trained using `fitcnet` command with `OptimizeHyperparameters` set to all. It explores: choice of activation function (`none`, `tanh`, `sigmoid`, `relu`), standardization (`true`, `false`), learning rate, layer weights initializer (`he`, `glorot`), layer biases initializer (`zeros`, `ones`) and layer sizes (for a given problem with 80 normal/control training cases and 188 training AMD/disease case it selects up to `3 layers`). The configuration is optimized for each partition separately. As can be seen in Table I, training GM-NSC is ten to hundred times faster than training the SVM classifier, and more the three orders of magnitude faster than training the NN classifier. Despite fully optimized 3-layers architecture NN classifiers yield smaller accuracy than GM-NSC. Likewise, GM-NSC yields accuracy comparable with accuracy of ResNet18 classifier, especially for smaller CRs. We contribute that to the small training set.

### 2) *Glaucoma Grading*

Corresponding dataset contains 151 normal cases, 77 cases of early glaucoma and 72 cases of advanced glaucoma [26]. Thus, the problems associated with the small sample scenario are expected to be even more pronounced in this case, especially in early vs. advanced-classes scenario. We evaluated GM-NSC, SVM, NN and deep CNN (DCNN) as a baseline on four scenarios: normal vs. early; normal vs. advanced; early vs. advanced; and normal vs. early vs. advanced. While GM-NSC, SVM and NN were evaluated on 100 random partitions, DCNN was evaluated on 30 random partitions only. GM-NSC, SVM and NN classifiers were applied to feature vector extracted by MERA-tensor network with CR $\in$ {4, 7, 10, 20, 40}. Corresponding ambient space dimension is $D \in$ {58564, 18769, 9025, 2045, 441}. For each value of CR, we cross-validated subspace dimension $d$ on 100 random partitions in terms of accuracy, F1 score, precision and recall. Fig. 5 shows accuracy vs. subspace dimension of four classifiers for CR=10.

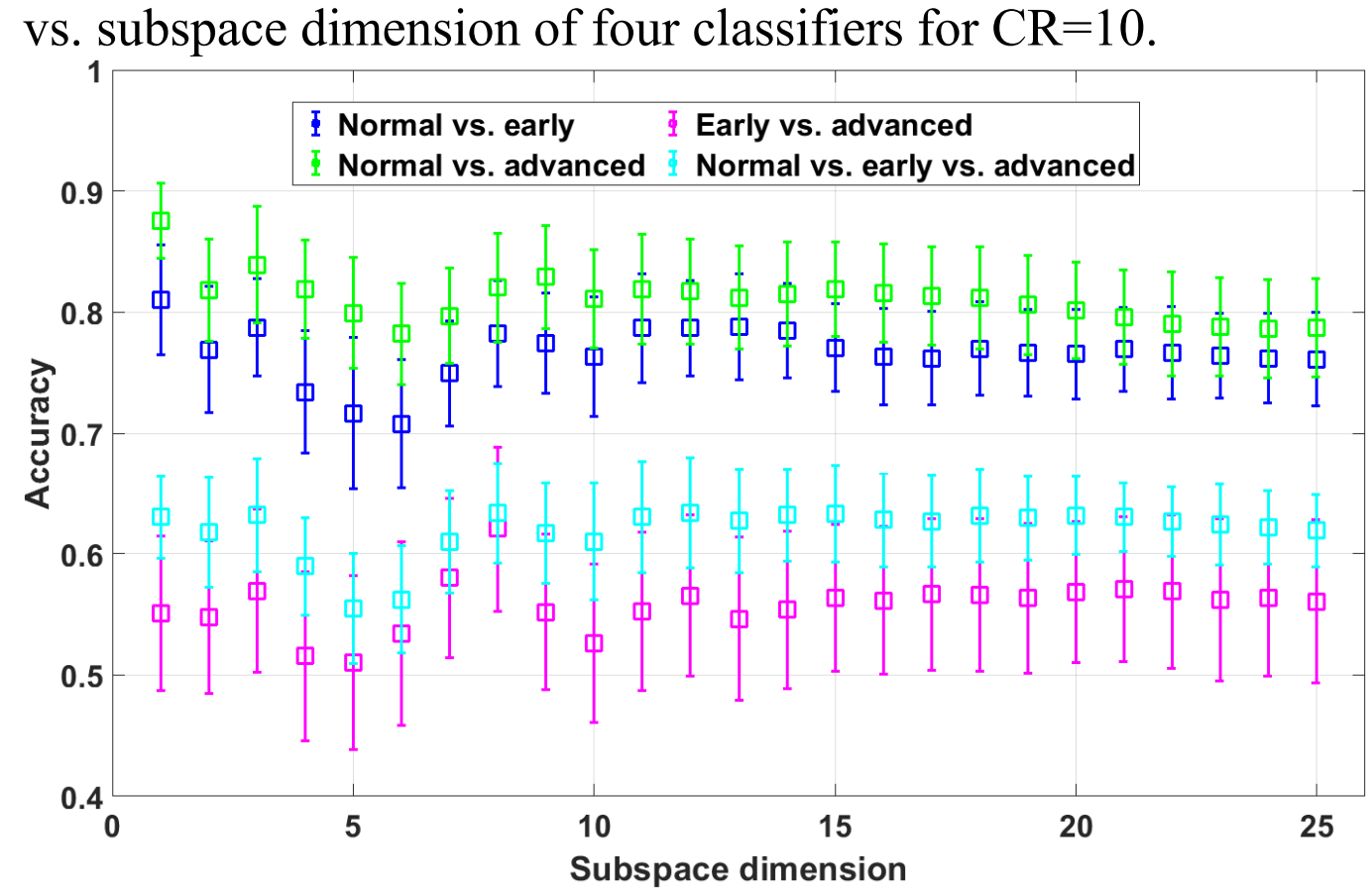

Fig. 5. Accuracy vs. subspace dimension $d$ for CR=10 for four GM-NSC classifiers.

As can be seen for cases normal vs. advanced and normal vs. early the highest accuracy occurs at $d$=1. For case early vs. advanced it occurs at $d$=8 and for case normal vs. early vs. advanced it occurs at $d$=12. Equivalent results (not shown here) are obtained for other three metrics. In all cases high dimensional MERA feature vectors have low-dimensional representations which constitutes basis for computationally efficient GM-NSC. Regarding SVM classifier, we conducted ablation study on MERA extracted feature vector for CR=10 for case normal vs. advanced. As in the case of normal vs. AMD dataset, linear SVM yielded the best result suggesting that MERA extracted feature vectors are discriminative and separable by hyperplanes. Thus, we used linear SVM in the rest of the analysis. We present results of GM-NSC, SVM and NN classifiers on MERA-feature vectors for CR$\in$ {4, 7, 10, 20, 40} in Tables II, III, IV and V for four scenarios. Direct comparison between GM-NSC and SVM classifiers reveals that the former mostly achieves better performance, sometimes even significantly better (early vs. advanced classifiers) with roughly two orders of magnitude lower computational complexity. Similar conclusions are drawn in comparative analysis between GM-NSC and NN classifiers. DCNN classifier yielded the highest performance with the exception of the early vs. advanced case, but also with significantly higher computational complexity. Due to very small training set (54/50), in early vs. advanced scenario DCNN classifier yielded worse result than GM-NSC classifier.

TABLE II
RESULTS OF OPTIMIZED SVM, NN AND GM-NSC CLASSIFIERS ON MERA-EXTRACTED FEATURE VECTORS FOR SELECTED CR. SCENARIO NORMAL VS. EARLY. THE BEST RESULT IS IN BOLD.

| CR | Classifier | ACC [%] | CPU [s] |
|---|---|---|---|
| 4 | NN | 80.18±4.28 | 420.02 |
| | SVM | 80.00±4.57 | 1850.1 |
| | **GM-NSC** | **81.01±3.60** | **10.38** |
| 7 | NN | 79.78±5.20 | 152.64 |
| | SVM | 80.71±4.82 | 227.67 |
| | **GM-NSC** | **81.94±4.03** | **1.107** |
| 10 | NN | 81.91±4.22 | 986.67 |
| | SVM | **82.35±4.72** | 89.13 |
| | GM-NSC | 81.04±4.54 | **0.300** |
| 20 | NN | 75.82±4.88 | 692.70 |
| | SVM | 78.30±4.52 | 27.52 |
| | **GM-NSC** | **81.44±3.95** | **0.053** |
| 40 | NN | 74.13±6.58 | 254.82 |
| | SVM | 76.74±6.70 | 23.12 |
| | **GM-NSC** | **82.07±3.59** | **0.005** |
| | DCNN | 90.83±4.52 | 5222 |

TABLE III
RESULTS OF OPTIMIZED SVM, NN AND GM-NSC CLASSIFIERS ON MERA-EXTRACTED FEATURE VECTORS FOR SELECTED CR. SCENARIO NORMAL VS. ADVANCED. THE BEST RESULT IS IN BOLD.

| CR | Classifier | ACC [%] | CPU [s] |
|---|---|---|---|
| 4 | NN | 81.44±5.45 | 358.40 |
| | SVM | 84.42±3.39 | 1867.2 |
| | **GM-NSC** | **84.88±3.32** | **10.83** |
| 7 | NN | 82.54±4.86 | 3882.4 |
| | SVM | 84.08±4.11 | 219.61 |
| | **GM-NSC** | **86.24±3.42** | **1.21** |
| 10 | NN | 82.70±5.57 | 1128.1 |
| | SVM | **89.79±4.00** | 84.22 |
| | GM-NSC | 87.52±3.11 | **0.33** |
| 20 | NN | 79.44±6.55 | 744.23 |
| | SVM | **84.38±4.91** | 24.84 |
| | GM-NSC | 82.71±4.45 | **0.03** |
| 40 | NN | 78.32±7.55 | 244.73 |
| | SVM | 82.32±6.98 | 22.57 |
| | **GM-NSC** | **83.70±4.15** | **0.005** |
| | DCNN | 95.33±2.44 | 5348 |

TABLE IV
RESULTS OF OPTIMIZED SVM, NN AND GM-NSC CLASSIFIERS ON MERA-EXTRACTED FEATURE VECTORS FOR SELECTED CR. SCENARIO EARLY VS. ADVANCED. THE BEST RESULT IS IN BOLD.

| CR | Classifier | ACC [%] | CPU [s] |
|---|---|---|---|
| 4 | NN | 51.52±4.82 | 70.10 |
| | SVM | 49.59±5.71 | 1799.2 |
| | **GM-NSC** | **58.84±6.15** | **13.16** |
| 7 | NN | 49.43±4.08 | 1730.7 |
| | SVM | 50.97±6.94 | 221.97 |
| | **GM-NSC** | **60.20±6.66** | **2.24** |
| 10 | NN | 51.54±6.55 | 1078.8 |
| | SVM | 53.30±7.94 | 86.58 |
| | **GM-NSC** | **62.05±6.80** | **0.64** |
| 20 | NN | 51.61±5.09 | 402.48 |
| | SVM | 49.20±7.07 | 26.36 |
| | **GM-NSC** | **62.70±5.61** | **0.05** |
| 40 | NN | 51.00±5.65 | 165.78 |
| | SVM | 50.80±8.27 | 26.52 |
| | **GM-NSC** | **59.21±6.56** | **0.01** |
| | DCNN | 57.90±6.27 | 3655 |

TABLE V
RESULTS OF OPTIMIZED SVM, NN AND GM-NSC CLASSIFIERS ON MERA-EXTRACTED FEATURE VECTORS FOR SELECTED CR SCENARIO NORMAL VS. EARLY VS. ADVANCED. THE BEST RESULT IS IN BOLD.

| CR | Classifier | ACC [%] | CPU [s] |
|---|---|---|---|
| 4 | NN | 60.28±4.56 | 96.85 |
| | SVM | 61.62±4.62 | 2169.6 |
| | **GM-NSC** | **63.45±4.25** | **21.28** |
| 7 | NN | 59.44±4.94 | 4156.4 |
| | SVM | **62.87±4.21** | 384.0 |
| | GM-NSC | 62.82±4.54 | **3.52** |
| 10 | NN | 60.07±5.01 | 1499.0 |
| | SVM | **66.84±4.55** | 160.9 |
| | GM-NSC | 63.39±4.59 | **1.04** |
| 20 | NN | 56.02±7.44 | 807.46 |
| | SVM | 61.30±5.61 | 98.11 |
| | **GM-NSC** | **63.61±5.15** | **0.08** |
| 40 | NN | 55.80±7.37 | 284.35 |
| | SVM | 58.86±5.66 | 90.79 |
| | **GM-NSC** | **63.06±4.57** | **0.008** |
| | DCNN | 75.63±4.27 | 6818 |

## VI. CONCLUSION

In this paper we proposed novel structure-aware model-driven compression ratio constrained low-rank MERA representation learning for compression, denoising and feature extraction from 3D OCT image volumes. Unlike learning-based approaches, the proposed framework is training-free and processes each OCT volume independently, making it well suited for applications with limited training data and maintaining high interpretability. Furthermore, we also proposed Grasmann-manifold nearest subspace classifier (GM-NSC) for diagnosis of retinal diseases enabling good generalization performance in small-sample settings scenario. On images acquired on two different scanners proposed method demonstrates enhanced visual quality and reduced retinal layer segmentation error at compression ratio as high as 40, especially on images acquired on low-quality scanner. We demonstrated on two clinical datasets that MERA-extracted high-dimensional feature vectors enjoy low-dimensional representation what enables GM-NSC based high-quality diagnosis of retinal diseases in small-sample clinical settings.

## REFERENCES

[1] W. Drexler and J. G. Fujimoto, *Optical Coherence Tomography: Technology and Applications*. Berlin, Germany: Springer, 2015.

[2] J. M. Schmitt, S. H. Xiang, and K. M. Yung, "Speckle in optical coherence tomography: an overview," *J. Biomed Opt.*, vol. 4, no. 1, pp. 95-105, 1999.

[3] D. Huang D, et al., "Optical coherence tomography," *Science*, vol. 254, pp. 1178-1181, 1991.

[4] P. G. Daneshmand and H. Rabbani, "Tensor Ring Decomposition Guided Dictionary Learning for OCT Image Denoising," *IEEE Trans. Med. Imag.*, vol. 43, no. 7, pp. 2547-2562, 2024.

[5] M. Gende, et al., "Automatic Segmentation of Retinal Layers in Multiple Neurodegenerative Disorder Scenarios," *IEEE J. Biomed. Health Inf.*, vol. 27, no. 12, pp. 5483-5494, 2023.

[6] J. De Fauw, et al., "Clinically applicable deep learning for diagnosis and referral in retinal disease," *Nature Med.*, vol. 24, no. 9, pp. 1342-1350, 2018.

[7] M. Elad and M. Aharon, “Image denoising via sparse and redundant representations over learned dictionaries,” *IEEE Trans. Image Process*., vol. 15, no. 12, pp. 3736–3745, 2006.

[8] W. Dong, L. Zhang, G. Shi, and X. Li, “Nonlocally centralized sparse representation for image restoration,” *IEEE Trans. Image Process*., vol. 22, no. 4, pp. 1620–1630, 2013.

[9] X. Gong, W. Chen, J. Chen, and B. Ai, “Tensor denoising using low-rank tensor train decomposition,” *IEEE Signal Process. Lett.,* vol. 27, pp. 1685–1689, 2020.

[10] I. Kopriva, et al., “Low Tensor Train and Low Multilinear Rank Approximations of 3D Tensors for Compression and De-Speckling of Optical Coherence Tomography Images,” *Phys. Med. Biol.*, vol. 68, no. 12, 2023, Art. no. 125002.

[11] L. I. Rudin, S. Osher, and E. Fatemi, “Nonlinear total variation based noise removal algorithms,” *Phys. D, Nonlinear Phenomena*, vol. 60, pp. 259–268, Nov. 1992.

[12] K. Zhang, W. Zuo, Y. Chen, D. Meng, and L. Zhang, “Beyond a Gaussian denoiser: Residual learning of deep CNN for image denoising,” *IEEE Trans. Image Process*., vol. 26, no. 7, pp. 3142–3155, 2017.

[13] K. Zhang, W. Zuo, and L. Zhang, “FFDNet: Toward a fast and flexible solution for CNN-based image denoising,” *IEEE Trans. Imag. Process*., vol. 27, no. 9, pp. 4608–4622, 2018.

[14] L. Fang, et al., “3-D Adaptive Sparsity Based Image Compression With Applications to Optical Coherence Tomography,” *IEEE Trans. Med. Imag.*, vol. 34, no.6, pp. 1306-1320, 2015.

[15] R. Rasti, H. Rabbani, A. Mehridenhavi, and F. Hajizadeh, “Macular OCT classification using a multi-scale convolutional neural network ensemble,” *IEEE Trans. Med. Imag.*, vol. 37, no. 5, pp. 1024-1034, 2018.

[16] L. Ju, et al., “Hierarchical Knowledge Guided Learning for Real-World Retinal Disease Recognition,” *IEEE Trans. Med. Imag.*, vol. 43, no. 1, pp. 335-350, 2024.

[17] S. Panayides, et al., “AI in medical imaging informatics: Current challenges and future directions,” *IEEE J. Biomed. Health Inform.*, vol. 24, no. 7, pp. 1837-1857, 2020.

[18] X. Chen *et al.*, “Recent advances and clinical applications of deep learning in medical image analysis,” *Med. Image Anal.*, vol. 79, 2022, Art. no. 102444.

[19] D. S. Kermany *et al*., “Identifying medical diagnoses and treatable diseases by image-based deep learning,” *Cell*, vol. 172, no. 5, pp.1122-1131.e9, Feb. 2018.

[20] S. Chen *et al.*, “FIT-Net: Feature interaction transformer network for pathologic myopia diagnosis,” *IEEE Trans. Med. Imag.*, vol. 42, no. 9, pp. 2524-2538, 2023.

[21] G. Vidal, “Class of quantum many-body states that can be efficiently simulated,” *Phys. Rev. Lett.*, vol. 101, no. 11,2008. Art. no. 110501.

[22] K. Batselier, A. Cichocki, and N. Wong, “MERACLE: Contrastive layer-wise conversion of a tensor train into mera,” *Commun. Appl. Math. Comput.*, vol. 3, no. 2., pp. 257-279, 2021.

[23] A. Jukić, I. Kopriva, and A. Cichocki, “Tensor decomposition-based feature extraction for noninvasive diagnosis of melanoma from the clinical color images,” *Biomed. Sig. Proc. Cont.*, vol. 8, no. 6, pp. 755-763, 2013.

[24] M. Udell and A. Townsend, “Why are big data matrices approximately low rank?” *SIAM J. Math. Data Sci.*, vol. 1, no. 1., pp. 144-160, 2019.

[25] S. Farsiu, et al., “Quantitative Classification of Eyes with and without intermediate Age-related Macular Degeneration Using Optical Coherence Tomography,” *Ophtalmology*, vol. 121, no. 1, pp. 162-172, 2014.

[26] J. Wu et al., “GAMMA challenge: Glaucoma grAding from Multi-Modality imAges,” *Med. Imag. Anal.*, vol. 90, 2023, Art. no. 102938.

[27] Z. Amini, R. Kafieh, and H. Rabbani, “Speckle noise reduction and enhancement for OCT images,” in Retinal Optical Coherence Tomography Image Analysis. Berlin, Germany: Springer, 2019, pp. 39–72.

[28] M. Szkulmowski et al., “Efficient Reduction of Speckle Noise in Optical Coherence Tomography,” *Opt. Exp.*, vol. 20, no. 2, pp. 1337-1359, 2012.

[29] A. Ozcan et al., “Speckle reduction in optical coherence tomography using digital filtering,” *J. Opt. Soc. Am. A*, vol. 24, no. 7, pp. 1901-1910, 2007

[30] J.-S. Lee, “Digital image enhancement and noise filtering by use of local statistics,” *IEEE Trans. Pattern Anal. Mach. Intell*., vol. PAMI-2, no. 2, pp. 165–168, 1980.

[31] L. Fang, S. Li, Q. Nie, J. A. Izatt, C. A. Toth, and S. Farsiu, “Sparsity based denoising of spectral domain optical coherence tomography images,” *Biomed. Opt. Exp*., vol. 3, no. 5, pp. 927–942, 201. Fang et al., “Fast acquisition and reconstruction of optical coherence tomography images via sparse representation,” *IEEE Trans. Med. Imag*., vol. 32, no. 11, pp. 2034–2049, 2013.

[32] I. Kopriva, F. Shi, and X. Chen, “Enhanced low-rank + sparsity decomposition for speckle reduction in optical coherence tomography,” *J. Biomed. Opt*., vol. 21, no. 7, 2016, Art. no. 076008.

[33] D. Thapa, K. Raahemifar, and V. Lakshminarayanan, “Reduction of speckle noise from optical coherence tomography images using multiframe weighted nuclear norm minimization method,” *J. Modern Opt*., vol. 62, no. 21, pp. 1856–1864, Dec. 2015.

[34] P. G. Daneshmand, A. Mehridehnavi, and H. Rabbani, “Reconstruction of optical coherence tomography images using mixed low rank approximation and second order tensor based total variation method, *IEEE Trans. Med. Imag*., vol. 40, no. 3, pp. 865–878, 2021.

[35] L. R. Tucker, “Some mathematical notes on three-mode factor analysis,“ *Psychometrika*, vol. 31, pp. 279-311, 1966.

[36] I. V. Oseledets, “Tensor-Train Decomposition,“ *SIAM J. Sci. Comp.*, vol. 33, pp. 2295-2317, 2011

[37] Q. Zhao, G. Zhou, S. Xie, L. Zhang, and A. Cichocki, “Tensor ring decomposition,” 2016, arXiv:1606.05535.

[38] Z. Long, C. Zhu, J. Chen, Z. Li, Y. Ren, and Y. Liu, "Multi-View MERA Subspace Clustering," *IEEE Trans. Multimed.*, vol. 26, pp. 3102-3112, 2024.

[39] Q. Xiao, L. Zhao, S. Chen, and X. Li, "Hyperspectral Anomaly Detection via MERA Decomposition and Enhanced Total Variation Regularization," *IEEE Trans. Geosc. Remote Sens.*, vol. 62, 2024, Art. no. 5514919.

[40] Y. Ma et al., “Speckle noise reduction in optical coherence tomography images based on edge sensitive cGAN,” *Biomed. Opt. Exp*., vol. 9, no. 11, pp. 5129–5146, 2018.

[41] F. Shi et al., “DeSpecNet: a CNN-based method for speckle reduction in retinal optical coherence tomography images,” *Phys. Med. Biol.* 64, 2019. Art. no. 175010.

[42] M. Wang M et al., “Semi-Supervised Capsule cGAN for Speckle Noise Reduction in Retinal OCT,” *IEEE Trans. Med. Imag.*, vol. 40, no. 4, pp. 1168-1183, 2021.

[43] T. G. Kolda and B. W. Bader, “Tensor decompositions and applications,” *SIAM Review*, vol. 51, pp. 455–500, 2009.

[44] B. Wang, Y. Hu, J. Gao, Y. Sun, F. Ju, and B. Yin, “Adaptive fusion of heterogeneous manifolds for subspace clustering,” *IEEE Transact. Neural Networks Learn. Syst.,* vol. 32, no. 8, pp. 3484–3497, 2021.

[45] I. Kopriva, D. Sitnik, L. -I. Dion-Bertrans, M. Milković Periša, A. Pačić, M. Hadžija, M. Popović Hadžija, “A hyperspectral imaging dataset and Grassmann manifold methods for intraoperative pixel-wise classification of metastatic colon cancer in the liver,” *Comp. Biol. Med*., vol. 196, 2025, Art. no. 110841.

[46] Y.M. Lu and M.N. Do, “A theory of sampling signals from a union of subspaces,” *IEEE Trans. Signal Process.*, vol. 56, no. 6, pp. 2334–2345,2008.

[47] J. Wright and Y. Ma, *High-Dimensional Data Analysis with Low-Dimensional Models – Principles, Computation and Applications.* Cambridge, United Kingdom: Cambridge University Press, 2022.

[48] P.A. Absil, R.E. Mahony, R. Sepulchre, *Optimization Algorithms on Matrix Manifolds*, Princeton University Press, 2008.

[49] P.A. Traganitis and G.B. Giannakis, “Sketched subspace clustering,” *IEEE Trans. Signal Process*., vol. 66, pp. 1663–1675, 2018.

[50] C. Davis and W. M. Kahan, “The rotation of eigenvectors by a perturbation. III,” *SIAM. J. Num. Anal.,* vol. 7, no. 1, pp. 1-45, 1970.

[51] Y. Yu, T. Wang, and R. J. Samworth, “A useful variant of the Davis-Kahan theorem for statisticians,” Biometrika, vol. 102, no. 2, pp. 315-323, 2015.

[52] X. Chen et al. , “Quantitative analysis of retinal layers' optical intensities on 3D optical coherence tomography,” *Invest. Ophthalmol. Vis. Sci.*, vol. 54, no. 10, pp. 6846-6851, 2013.

[53] Y. B. Zheng et al., " Fully-Connected Tensor Network Decomposition and Its Application to Higher-Order Tensor Completion," in *Proc. of the AAAI-21*, vol. 35, No. 12, February 2-9, 2021.

[54] F. Shi et al., “Automated 3-D Retinal Layer Segmentation of Macular Optical Coherence Tomography Images with Serous Pigment Epithelial Detachments,” *IEEE Trans. Med. Imag.*, vol. 34, no. 2, pp. 441-452, 2015.

[55] M. Hänsel et al., “Retinal layer thickness as a potential marker for future cardiovascular events – a UK Biobank cohort using optical coherence tomography,” *J. Neurol.* Vol. 273, 2026, Art. no. 662.